\pdfoutput=1
\documentclass{article}

\usepackage[nonatbib, preprint]{neurips_2021}

\usepackage[utf8]{inputenc} 
\usepackage[T1]{fontenc}    
\usepackage{hyperref}       
\usepackage{url}            
\usepackage{booktabs}       
\usepackage{amsfonts}       
\usepackage{nicefrac}       
\usepackage{microtype}      
\usepackage{xcolor}         
\usepackage{graphicx}
\usepackage{subcaption}
\usepackage{makecell}

\title{Beyond the Gates of Euclidean Space: Temporal - Discrimination - Fusions and Attention-based Graph Neural Network for Human Activity Recognition}

\author{%
  Nafees Ahmad*\\
  Department of Computer Science and Engineering \\
  The Chinese University of Hong Kong \\
  Shatin, N.T., Hong Kong SAR \\
  \texttt{nafees@link.cuhk.edu.hk}
  \And
   Savio Ho-Chit Chow \\
   Department of Computer Science and Engineering \\
  The Chinese University of Hong Kong \\
  Shatin, N.T., Hong Kong SAR \\
  \texttt{saviochowhochit@link.cuhk.edu.hk}
   \AND
   Ho-fung Leung\\
   The Chinese University of Hong Kong \\
   Shatin, N.T., Hong Kong SAR \\
  \texttt{lhf@cuhk.edu.hk}
}

\begin{document}

\maketitle

\long\def\/*#1*/{}

\begin{abstract}

Human activity recognition (HAR) through wearable devices has received much interest due to its numerous applications in fitness tracking, wellness screening, and supported living. As a result, we have seen a great deal of work in this field. Traditional deep learning (DL) has set a state of the art performance for HAR domain. However, it ignores the data's structure and the association between consecutive time stamps. To address this constraint, we offer an approach based on Graph Neural Networks (GNNs) for structuring the input representation and exploiting the relations among the samples.  However, even when using a simple graph convolution network to eliminate this shortage, there are still several limiting factors, such as inter-class activities issues, skewed class distribution, and a lack of consideration for sensor data priority, all of which harm the HAR model's performance. To improve the current HAR model's performance, we investigate novel possibilities within the framework of graph structure to achieve highly discriminated and rich activity features. We propose a model for (1) time-series-graph module that converts raw data from HAR dataset into graphs; (2) Graph Convolutional Neural Networks (GCNs) to discover local dependencies and correlations between neighboring nodes; and (3) self-attention GNN encoder to identify sensors interactions and data priorities. To the best of our knowledge, this is the first work for HAR, which introduces a GNN-based approach that incorporates both the GCN and the attention mechanism. By employing a uniform evaluation method, our framework significantly improves the performance on hospital patient's activities dataset comparatively considered other state of the art baseline methods.


\end{abstract}

\section{Introduction}
\label{intro}

Recent availability of compact and mobile smart devices has transformed society in terms of connectivity and mode of interaction. With the miniaturization of smart devices and increasing number of sensors available on these devices, daily activity tracking has grown in popularity. Human activity recognition (HAR) is an active research domain which involves integrating data generated from multiple sensor types such as accelerometers, gyroscopes, etc. to detect the status or activity an individual is undergoing.

\subsection{Problem Statement and Motivation}

Traditional deep learning (DL) has established a benchmark for HAR performance; however, extensive complex activities, identical patterns for different activities, noise, and diverse representations of a single activity significantly impair the performance of current HAR models. Existing HAR methods use raw time-series data as the model’s input. However, this raw data is unstructured and lacks any information about the relationships between representations of different samples. Thus, we hypothesize that converging unstructured data into structured data can contribute to modelling these diverse, challenging activities.

That is why, in order to improve the performance of the HAR model, we found a key opportunity to move activity representations from Euclidean to non-Euclidean space. This enables the model to be trained end-to-end on structured data rather than unstructured input.  Thus, the model is capable of representing each time stamp using both its own and nearby time stamps features. To leverage structural representation and boost the model's ability to learn more discriminative features, we suggest an approach based on Graph Neural Networks (GNNs). We reveal the following dimensions within the GNN umbrella.

\begin{enumerate}

    \item The nearby acceleration readings in human activity are likely to be correlated. This imposes that the model captures the local dependencies to get better discriminative representations.

    \item Multiple sensors are used to collect data for HAR (attached to the different body positions). Generally, some sensors contribute significantly more to ongoing activity than others (For example, in drinking activity, right-hand sensors are used to involve more than left-hand and foot sensors). This gap suggests that we should leverage sensor interactions and prioritize sensors based on their relevance to the current activity.
    
\end{enumerate}

\subsection{Contributions}

\begin{enumerate}

    \item To shape the input data for the GNN, we design a time-series-graph module that converts raw data into graphs. This module explicitly captures structural information sample by sample through connections (edges). The edges also directly connect samples separated in time with $O(1)$ complexity, making modelling easier. This method converts the unstructured time series to a structured graph with rich semantics, which benefits the HAR task processing (\textbf{Section~\ref{tg}, Figure~\ref{fig:ga}}).
    
    \item We propose to use Graph Convolutional Neural Networks (GCNs) to discover local dependencies and correlations between neighboring nodes (timestamps) (\textbf{Section~\ref{gcn}}). GCN is based on node-to-node message passing techniques to learn both local and global time stamp dependencies.
    
    \item We propose the self-attention GNN encoder module for identifying sensor interactions and prioritizing sensors relevant to the current activity. This encoder module employs self-attention~\cite{DBLP:conf/icml/LeeLK19} to learn the interactions between sensors and exploit the capabilities of different sensors. Additionally, this method computes the correlation between the sensors in order to produce self-attention maps, which enrich the representations for subsequent tasks (\textbf{Section~\ref{self-attention}}).

    \item Using standard evaluation protocols, we compare our method performance to the most recent state-of-the-art bench marks HAR research to illustrate the effectiveness of our work and its improvement. In addition, since we could not find any previous work for HAR in the GNN domain, to this end, we offer the Recurrent attention Graph neural Network (RAGNN) technique as a baseline (\textbf{Section~\ref{rag}}). It leverages to capture the relationships between sequential time stamps and learns contextually relevant timestamps. We inspire RGNN~\cite{DBLP:journals/tsp/RuizGR20,DBLP:conf/icassp/IoannidisMG19,DBLP:conf/iconip/SeoDVB18} and an attention-based approach to utilize this method. Recent works~\cite{DBLP:journals/corr/abs-2103-10760,DBLP:conf/cvpr/SiC0WT19,DBLP:conf/cvpr/KuenWW16} demonstrated that these methods could model sequences and enrich representations. Additionally, we involve the Convolutional Neural Network (CNN) experiments results to demonstrate the effectiveness of our work in comparison to the widely used CNN method.

    \item To the best of our knowledge, we are the first to introduce a GNN based method that includes the GCN and attention mechanism on HAR dataset. By employing a uniform evaluation method, our framework significantly improves the performance on hospital patient's activities dataset. Furthermore, we present the performance of our approach quantitatively (\textbf{Section~\ref{qun_r}}) and qualitatively to demonstrate its efficacy and generalizability (\textbf{Section~\ref{qul_r}}).

\end{enumerate}

\section{Problem Formulation}
This work aims to build an end-to-end graph neural network HAR model that consumes raw multidimension sensor data (captured through wearable devices) as input and accurately recognizes human activities. The data from multi-sensors can be represented using the matrix $X=\{x_1, x_2, x_3, ..., x_n\}\in\mathbb{R}^{D\times T}$, where $T$ represents time stamps and $D$ is the number of sensor channels across each timestamp. Matrix X is mapped to label Y from a predetermined set of human activities $(h_1, h_2, h_3, ..., h_n)$. In a more compact structure, we have supplied $\textbf{X}=\{X_1, X_2, X_3, ..., X_n\}\in\mathbb{R}^{N\times D\times T}$, $N$ represents the number of window segments, and the labels associated with this tensor are $\textbf{Y}=\{Y_1, Y_2, Y_3, ..., Y_n\}\in\mathbb{R}^{N}$. Our proposed method task is to learn and map the relationship between $\textbf{X}$ and $\textbf{Y}$.
\section{GNN Based Proposed HAR Framework}
\subsection{Data Preprocessing and window segmentation}

The adopted data set has down sampled to 33HZ (where required) to achieve temporal consistency between multiple data sets. The raw data from all sensors is then normalized dimension (sensor channel) wise to zero mean and unit variance.

\begin{figure}[t!]
\centering
\minipage{0.99\textwidth}
    \centering
    \includegraphics[width=5cm]{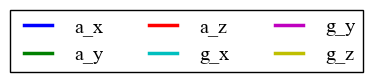}
\endminipage\hfill
\minipage{0.32\textwidth}
    \centering
    \includegraphics[width=\linewidth]{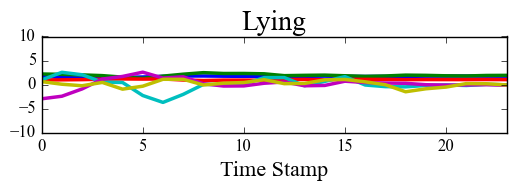}
\endminipage\hfill
\minipage{0.32\textwidth}
    \centering
    \includegraphics[width=\linewidth]{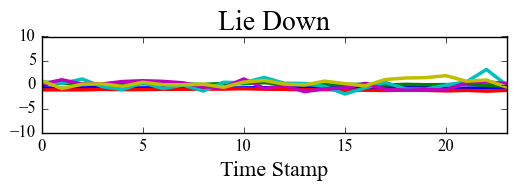}
\endminipage\hfill
\minipage{0.32\textwidth}
    \centering
    \includegraphics[width=\linewidth]{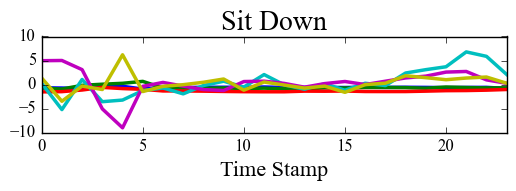}
\endminipage\hfill
\minipage{0.32\textwidth}
    \centering
\includegraphics[width=\linewidth]{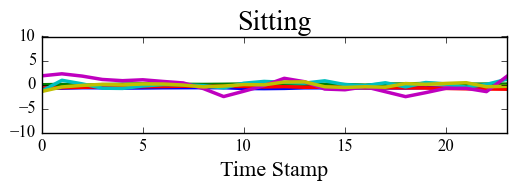}
\endminipage\hfill
\minipage{0.32\textwidth}
    \centering
    \includegraphics[width=\linewidth]{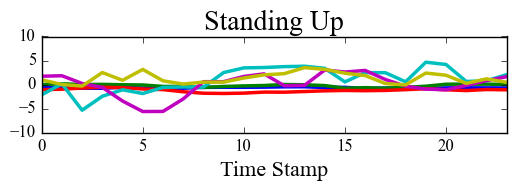}
\endminipage\hfill
\minipage{0.32\textwidth}
    \centering
    \includegraphics[width=\linewidth]{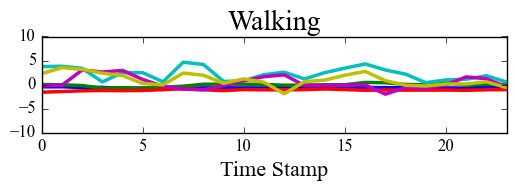}
\endminipage\hfill
\minipage{0.32\textwidth}
    \centering
    \includegraphics[width=\linewidth]{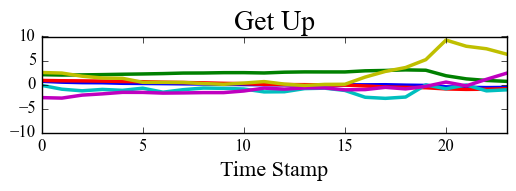}
\endminipage\hfill
\caption{\textit{Visualization of each activity's segmented data from the Hospital data set. The vertical axis represents the sensor's channel $(D=6)$ magnitude, while the horizontal axis depicts time stamps $(T=24)$.}}
\label{figRawData}
\end{figure}

We cannot directly insert all the raw data into the classification model. Before feeding the data to the machine learning model, it is necessary to create the data slices. Therefore, sliding window approach is used~\cite{10.1145/3090076} to make the data into partitions. We follow state-of-the-art studies~\cite{10.1145/3090076} to fix the segment duration, which employed $T=24$ with 50\% overlap. Intuitively, the compact format of the input segment corresponds to $D\times T$, means that if we have two sensors with three channels each, we will have a $6\times 24$ input matrix. On the other side, each timestamp had y (e.g., 24) labels attached; however, because we are now inputting 24 samples simultaneously, we give one label to the $D\times T$ input segment. The most frequently occurring labels (associated with time stamps) is nominated as single label to the corresponding segment. We visualize (see \textbf{Figure~\ref{figRawData}}) a segment data of each activity as an example for better understanding.

\subsection{Time-series-graph Module} \label{tg}

\begin{figure}[t!] 
\centering
\begin{subfigure}[t]{.32\linewidth}
    \minipage{\textwidth}
        \centering
        \includegraphics[width=\linewidth]{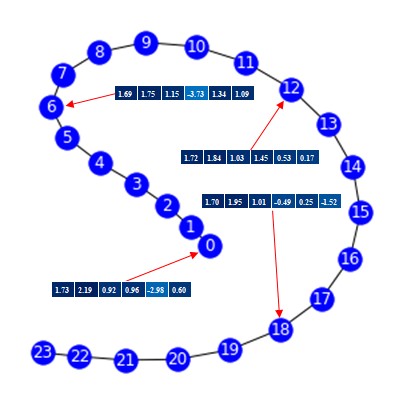}
        \caption{\textit{Lying}}
    \endminipage\hfill
\end{subfigure}
\begin{subfigure}[t]{.32\linewidth}
    \minipage{\textwidth}
        \centering
        \includegraphics[width=\linewidth]{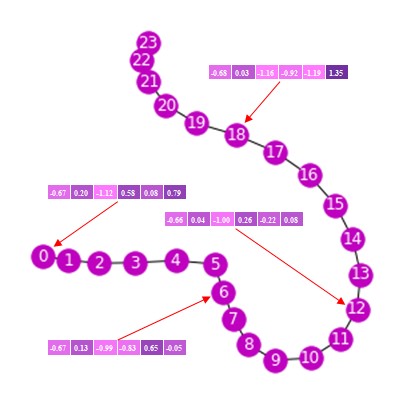}
        \caption{\textit{Lying Down}}
    \endminipage\hfill
\end{subfigure}
\begin{subfigure}[t]{.32\linewidth}
    \minipage{\textwidth}
        \centering
        \includegraphics[width=\linewidth]{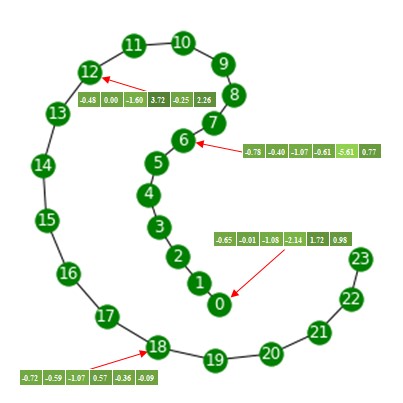}
        \caption{\textit{Standing Up}}
    \endminipage\hfill
\end{subfigure}
\begin{subfigure}[t]{.32\linewidth}
    \minipage{\textwidth}
        \centering
        \includegraphics[width=\linewidth]{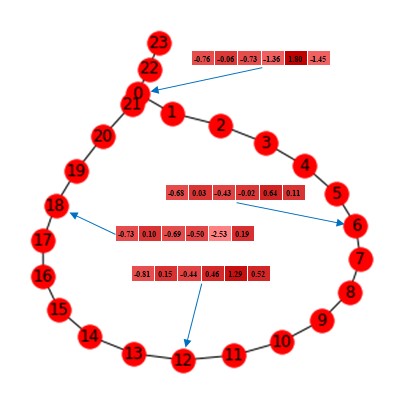}
        \caption{\textit{Sitting}}
    \endminipage\hfill
\end{subfigure}
\begin{subfigure}[t]{.32\linewidth}
    \minipage{\textwidth}
        \centering
        \includegraphics[width=\linewidth]{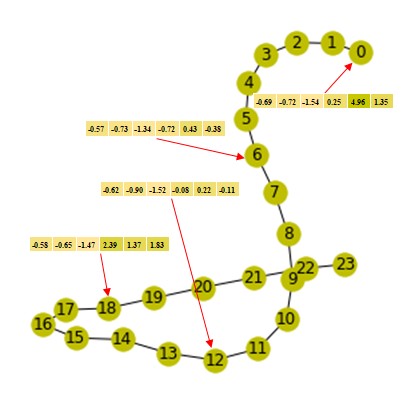}
        \caption{\textit{Sitting Down}}
    \endminipage\hfill
\end{subfigure}
\begin{subfigure}[t]{.32\linewidth}
    \minipage{\textwidth}
        \centering
        \includegraphics[width=\linewidth]{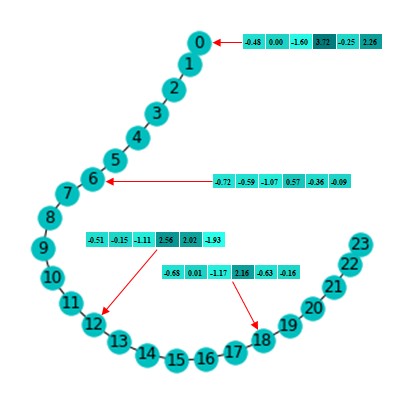}
        \caption{\textit{Walking}}
    \endminipage\hfill
\end{subfigure}
\begin{subfigure}[t]{.32\linewidth}
    \minipage{\textwidth}
        \centering
        \includegraphics[width=\linewidth]{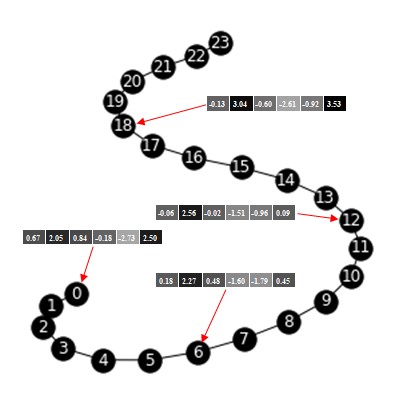}
        \caption{\textit{Getting Up}}
    \endminipage\hfill
\end{subfigure}
\caption{\textit{Graphs visualization of each activity's segmented data generated by our Time-series-graph module for hospital data set. The vector array in each graph indicates sensors channels features $(D_c)^{C=6}_{c=1}$ of the corresponding node.}}
\label{fig:ga}
\end{figure}

After segmenting the data, we pass $X\in\mathbb{R}^{D\times T}$ to the time-series graph encoder module, which converts the unstructured raw time series data to structured graph topological network data. Each timestamp $T$ is considered a node, while $D$ sensor channels are regarded as features of each node. Then, two consecutive timestamps (nodes) are connected using an undirected edge connection. The intuition for establishing undirect edge connections between nodes is that data about human activities are naturally in time series and that two successive time stamps are logically and contextually adjacent. Thus, the first- and last-time stamps nodes have degree 1, while the remaining nodes (timestamps) have degree 2. \textbf{Figure~\ref{fig:ga}} shows some of the graphs we constructed for each activity using raw segments.

We construct graphs $G\in\mathbb{R}^{D\times T}$ for each segment X separately and then feed them to a GNN model with a label $Y$ for the downstream task, this leads our problem is being graphs classification problem. In this manner, we build $\textbf{G}=\{G_1, G_2, G_3, ..., G_n\}\in\mathbb{R}^{N\times D\times T}$, where $N$ denotes the total number of graphs, $T$ is the total number of nodes in each graph, and $D$ denotes the total number of sensor channels across each node. $\textbf{Y}=\{Y_1, Y_2, Y_3, ..., Y_n\}\in\mathbb{R}^{N}$ are the labels for $\textbf{G}$. Therefore, our study's
objective is to precisely learn the relationship between $\textbf{G}$ and $\textbf{Y}$.
x

\begin{figure}[ht]
\centering
\includegraphics[width=13cm]{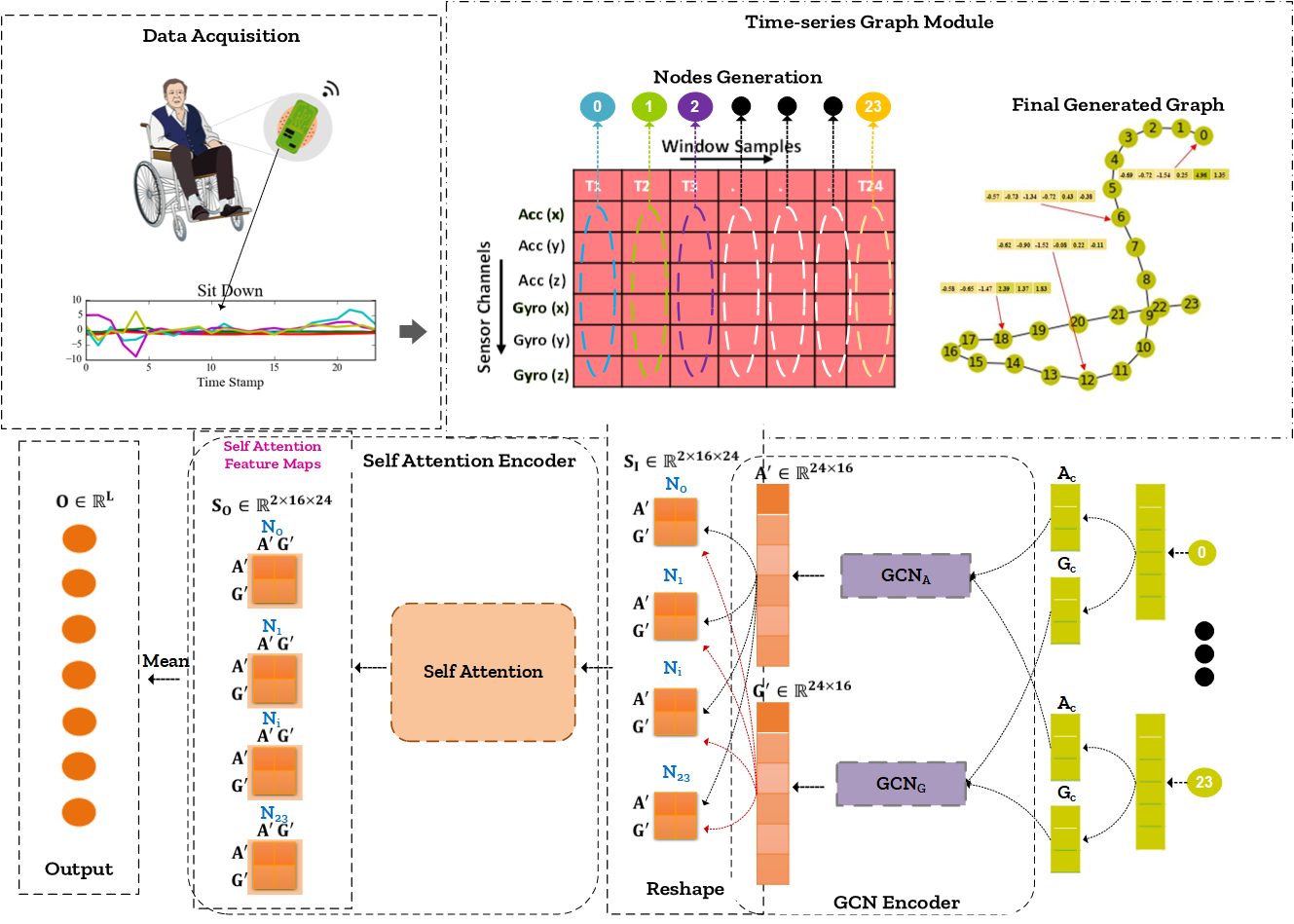}
\caption{Our approach.}
\label{fig:approach}
\end{figure}
\subsection{GCN Encoder}
\label{gcn}

The input of the model is a graph $G$ consisted of the adjacent matrix $A\in\mathbb{R}^{t\times t}$ and node feature matrix $H\in\mathbb{R}^{t\times n \times d}$, where $t$ is the number of nodes, $n$ is the number of sensors and $d$ is the number of features. Here we choose GCN \cite{DBLP:conf/iclr/KipfW17} to embed the graph, because it has proven to be robust in a wide range of tasks. Moreover, the degree of nodes is our graph is limited, so GCN has enough capacity to process features from neighbors and might enjoy additional regularization effect due to its simple architecture compared to other advanced GNN variants like GAN \cite{DBLP:conf/iclr/VelickovicCCRLB18}. Instead of feeding everything into a single GCN, we first split $G$ into multiple graphs $\mathbf{G}=\{G^1,\cdots,G^i,\cdots,G^n\}$. $G^i$ denotes a graph for the $i$-th sensor. $G^i$ has the same adjacent matrix $A$ as $G$, but its node feature matrix $H^i\in\mathbb{R}^{t\times d_i}$ only contains features from that $i$-th sensor, where $d_i$ is the number of features in the $i$-th sensor. Then we will use a separate GCN to process each $G^i$ independently, i.e., we will have $n$ GCNs for $n$ sensors.

GCN is a stack of layers and the formulation of the $l+1$ layer is shown as below:
\begin{equation}
    H^{(l+1)}=\sigma(D^{-\frac{1}{2}}AD^{-\frac{1}{2}}H^{(l)}W^{(l+1)})
\end{equation}
where $D$ and $A$ are the degree and adjacency matrices of the graph. $H^{(l)}$ is the output of $l$ layer and $W^{(l+1)}$ is the weight of $l+1$ layer. $\sigma$ is the ReLU activation function.

The output of our GCNs will be a set of embedded node matrices $\mathbf{\hat{H}}=\{\hat{H^1},\cdots,\hat{H^i},\cdots,\hat{H^n}\}$ and $\hat{H^i}\in\mathbb{R}^{t\times \hat{d}}$, where $\hat{d}$ is the hidden size.

\subsection{Self-Attention GCN Encoder}
\label{self-attention}

A reliable activity recognition model always relies on multiple sensor data obtained from the user's different body positions. Each sensor provides a unique perspective on the current undergoing activity. For instance, in a scenario of drinking activity, the Right-Hand sensors (Upper and Lower Hand Sensors) will capture significantly more information than other body sensors. However, general HAR methods incorporate data from all sensors equally into the model to recognize the activities, which causes the model to be misled during the training. 

Furthermore, it is also important to examine the contributions of other sensors based on their level of interaction with key sensors. Therefore, categorizing sensors based on their relevance to activity and interactions with one another is highly essential. Hence, accordingly, we develop an end-to-end trainable self-attention encoder module that accepts GCN feature maps as input for each node, captures the highly active sensor feature map, and learns the interaction among each pair of sensors. Consequently, in this manner, this module provides rich aggregated representation for later classification.

\paragraph{Inter-Sensor Self-Attention}

Given the embedded node matrices $\mathbf{\hat{H}}$ from GCNs, the model iteratively apply the self-attention to aggregate information along the sensor dimension for nodes from different $G^i$ but at the same timestamp $j$.

Concretely, for a node from the $i$-th sensor at $j$-th timestamp, we extract its node features $\hat{H}^i_j$ from $\mathbf{\hat{H}}$ and stack all $\hat{H}^i_j$ from different sensors $i$ as $\mathbf{\hat{H}}_j=\{\hat{H}^1_j,\cdots,\hat{H}^i_j,\cdots,\hat{H}^n_j\}\in\mathbb{R}^{n\times \hat{d}}$. The self-attention will then be performed along the sensor dimension $n$ and it is defined as bellows \cite{NIPS2017_3f5ee243}:
\begin{eqnarray}
    \label{eqn:distribution}
    \hat{A}&=&\mathrm{softmax}\left(\frac{\mathbf{\hat{H}}_jW_q{(\mathbf{\hat{H}}_jW_k)}^T}{\sqrt{\hat{d}}}\right)\\
    \mathrm{Attention}(\mathbf{\hat{H}}_j)&=&\hat{A}\mathbf{\hat{H}}_jW_v
\end{eqnarray}
where $W_q, W_k, W_v \in \mathbb{R}^{\hat{d}\times\hat{d}}$ are learnable parameters. After applying self-attention to $t$ timestamps, the attention produce an output $\mathbf{\bar{H}}\in\mathbb{R}^{t\times n \times \hat{d}}$.

In Eq.~\ref{eqn:distribution}, $\hat{A}$ defines an attention distribution over sensors, where a pair of sensors with similar features will get a higher score. We can identify which sensor is important by using $\hat{A}$. If more sensors assign more probability mass to a specific sensor, then this specific sensor is more important.

\paragraph{Output Layer}

The output layer will do a mean pooling to the attention output $\mathbf{\bar{H}}$ along the timestamps dimension $n$, i.e., its output will have a shape of $n\times \hat{d}$. We do so because we want to obtain a graph-level representation for graph classification. Then we flatten the mean pooling output and pass it through a linear projection as well as a softmax layer to predict the probability of each possible activity.

\section{Baselines}

\subsection{Recurrent Attention Graph Neural Network (RAGNN)} \label{rag}

Time series nature of HAR data naturally demands methods which learn the context of the activity in consecutive timestamps. Previous works using traditional deep learning methods neglect the consecutive nature of HAR data which may be counter-intuitive to prediction on time series HAR data. Also, traditional HAR methods considers the contribution of all timestamps equally to the ongoing activity for a given segment. However, this may not be the case when we predict the activity from a segment, since some timestamps may be more crucial for prediction. Therefore it is necessary to learn the context in consecutive timestamps and filters out the unnecessary timestamps for further downstream tasks.

Taking recurrent neural networks as inspiration, each timestamp of an HAR data segment can be represented as the problem of feeding consecutive timestamps into a recurrent unit to learn hidden representation of sensor data, taking into consideration previous timestamps to extract the context of the entire HAR data segment. This captures the specific context between timestamps and their relationship in the entire segment, which models sequential data more naturally. Also inspired by graph attention mechanism, we can take into account the importance of each timestamp's towards final activity prediction of the entire segment represented as a graph by enriching timestamp representation. By considering the weight of each neighbourhood node (adjacent timestamp) during aggregation, we can take into account more important timestamps and dampen unnecessary timestamps for efficient downstream prediction.

Therefore, we propose a \textit{Recurrent Attention Graph Neural Network (RAGNN) encoder} module, first utilizing a LSTM-layer for each sensor of the input HAR graph segment to recurrently learn a contextually significant hidden representation of nodes, and then passing the learnt hidden representation to a graph attention network to aggregate the neighbouring nodes with respect to neighbourhood timestamp importance.

After representing time series HAR data in the form of a graph $G$, we have node features consisting of $d$ features (e.g. x, y and z-axis data from all sensors) coming from $n$ sensors (e.g. accelerometer and gyroscope) and $t$ number of nodes (each timestamp of an activity segment). Therefore, the input to the RAGNN consists of an adjacency matrix $A\in\mathbb{R}^{t\times t}$ and node feature matrix $H\in\mathbb{R}^{t\times d}$, where the node features can be further subdivided into features coming from each of the $n$ sensors, therefore $d$ equals $n$ * 3 if all sensors are tri-axial. $n$ LSTM layers will process the data from each sensor in parallel. The output is thus a node hidden representation $H\in\mathbb{R}^{t\times h*n}$, where $h$ is the hidden dimension size of a LSTM layer and $n$ is the total number of sensors.

The output hidden representation of nodes from the LSTM layers capture the contextual information of consecutive nodes per sensor, and will be fed in parallel per sensor to $n$ Graph Attention Networks (GAT). For a graph $G$ with adjacency matrix $A\in\mathbb{R}^{t\times t}$ and hidden representation matrix from LSTM layer $H\in\mathbb{R}^{t\times h*n}$, each sensor's learnt hidden representation will be the node feature matrix input to each GAT block.

The GAT layer is defined as follows:

\begin{eqnarray}
    \label{eqn:GAT}
    h_i^{(l+1)} = \sum_{j\in \mathcal{N}(i)} \alpha_{i,j} W^{(l)} h_j^{(l)}
\end{eqnarray}

where $h_i^{(l+1)}$ is the hidden representation of node $i$ in the $l$+1th layer, $\mathcal{N}(i)$ is the set of neighbors of node $i$, $\alpha_{ij}$ is the attention score bewteen node $i$ and node $j$, and $h_j^{(l)}$ is the hidden representation node $j$ (a neighbour of node $i$) in the $l$th layer. Specifically, $\alpha_{ij}$ is defined as follows:

\begin{eqnarray}
    \label{eqn:GAT_a_e}
    \alpha_{ij}^{l} &=& \mathrm{softmax_i} (e_{ij}^{l}) \\
    e_{ij}^{l} &=& \mathrm{LeakyReLU}\left(\vec{a}^T [W h_{i} \| W h_{j}]\right)
\end{eqnarray}

The attention weighting will assist in determining the significance of each timestamp. We anticipate that using this approach, the RAGNN module will encode relevant context and useful timestamps for downstream activity classification. The output of the final GAT layer will be flattened across all nodes and fed to seven neurons, and a SoftMax activation will compute the likelihood of each activity label.

\section{Dataset}
The dataset consisted of 12 hospitalized elderly patients wearing inertial sensors on their clothing to perform seven different categories of activities. The dataset was collected at a frequency of 10 Hz. For evaluation purposes, data from the first eight and next three participants were used for the testing and training phases. The rest were used for validation.

\section{Experiments, Evaluation and Results}
\subsection{Implementation Detail}

PyTorch and DGL libraries are employed for implementing the Self-Attention GCN encoder. The self-attention GCN encoder contains 5 GCN layers and one self-attention layer. The hidden size is 16 and the model is optimized by Adam optimizer with a learning rate of 0.01 and a batch size of 100. We train the model for 100 epochs and save one checkpoint at the end of each epoch. Then the checkpoint with the highest F1 score in the validation set is chosen to perform prediction on the test set.

For network visualization, Matplotlib is used to illustrate the attention distribution across sensors as well as the outputs of the second last layer in 2D surface. T-SNE algorithm is exploited to project the high-dimensional outputs for 2D visualization.

\subsection{Evaluation Method} The major challenge in recognizing human activity is that data collection is time-consuming. We can only have 10 to 15 individual data to train the model. However, everyone's activity pattern varies, even for the same activity, due to factors such as age, physical condition, and weight. Furthermore, even the same individual develops various patterns for the same action. For instance, a user walking in an office is somewhat different from a user walking in a park. To obtain an accurate HAR model, one solution is to collect data from the entire population for an unlimited number of activities and contexts (walking in a park versus walking in an office). Model on this huge data set can be  justifiable to evaluate by using the generic approach like dividing the data set into 70\% for training and 30\% for testing, but collecting that much huge data for HAR is entirely impossible. Therefore, a hold-out evaluation strategy ensures fair reporting, which divides the data into training and testing segments. In HAR, this referred as cross-subject and cross-run activity recognition. In cross-subject, the training data contains data from different subjects, whereas the testing data includes data from entirely separate subjects (completely unseen for the model). While in cross-run evaluation, data distribution is done based on different runs. For instance, in the first run, the user is instructed to walk in the morning and at home, whereas in the second run, the user is instructed to walk at school in the evening. Hence we considered hold-out evaluation method to report our model results.

For the training, segment-wise (24 samples with 12 overlapped) data is fed to the model. However, in testing, sample by sample (called sample wise evaluation) is predicted for establishing a more realistic setup.

Since the HAR data sets are highly skewed, utilizing an accuracy evaluation matrix does not constitute
fair reporting. This matrix gives preference to majority class data. To ensure fair reporting, the
F1-score evaluation matrix is used. We compute F1-score by using the following equation:
\begin{eqnarray}
    \label{eqn:f1score}
    \mathrm{F1} = 2 \times \frac{\mathrm{precision} \times \mathrm{recall}}{\mathrm{precision} + \mathrm{recall}}
\end{eqnarray}
Additionally, we report confusion matrices for proposed and baseline frameworks to demonstrate the model’ performance with respect to each activity.

\subsection{Comparison with the Cutting Edge Benchmark HAR Research}
To highlight the effectiveness of our proposed framework, we compared our findings to six state-of-the-art benchmark studies, as shown in Table 1. All of these approaches are trained on Euclidean space representations. All studies presented their findings in the form of f1-scores through the use of hold out evaluation and a sample wise prediction technique. Similar evaluation processes are followed for our study to assure fair judgment and objectivity.

Additionally, we developed two other approaches based on GNN and used their performance as a benchmark for the performance of our proposed framework. The reason for establishing these baselines is that GNN is a relatively new subject, with no work available to build baselines at the moment. That is why we ran trials utilizing GNN techniques that have already shown improved performance in computer vision and natural language processing.

\label{qun_r}
As shown in Table~\ref{tab:baseline}, our proposed method greatly outperformed baselines. 
All the baseline benchmark results are taken from the study~\cite{10.1145/3448083}. As can be seen, our approach outperformed all other methods by a wide margin. Attend and Discriminate~\cite{10.1145/3448083} have a reasonable performance margin compared to our work. One of the downsides of this method is that it improves performance by data augmentation.  

In the scenario of GNN based baselines, RAGNN with attention performs poorly compared to the classic CNN approach, so it should not be considered for this area. By contrast, a single GCN outperformed a traditional CNN by 3.1\%. It is a worthwhile enhancement to the complex, noisy data set.However, involving self-attention with GCNN (our proposed work), we are able to boost performance by more than 5\% dramatically. This demonstrates that taking into account all sensor data affects performance and that finding which sensor can best encode specific activity results in a huge rise in model performance.

\begin{table}[!htp]
\centering
\caption{\textit{Performance comparison between the proposed method and baselines.}}
\begin{tabular}{l|l}
\hline
\makecell[c]{\textbf{Model}}    & \makecell[c]{\textbf{F1-score}}  \\ \hline
LSTM Baseline~\cite{10.1145/3090076} & 62.7  \\ \hline
DeepConvLSTM~\cite{s16010115}  & 62.8  \\ \hline
Bi-directional LSTM~\cite{10.5555/3060832.3060835}   & 63.6  \\ \hline
Dense Labeling~\cite{YAO2018252}  & 62.9  \\ \hline
Attention Model~\cite{10.1145/3267242.3267287}   & 64.1  \\ \hline
Attend and Discriminate~\cite{10.1145/3448083}  & 66.6  \\ \hline
RAGNN    & 43  \\ \hline
GCN    & 63.1  \\ \hline
\textbf{Our Proposed Method (GCN + Self Attention)}    & \textbf{68.5}  \\ 
\end{tabular}
\label{tab:baseline}
\end{table}

\subsection{Qualitative results}
\label{qul_r}
\begin{figure}[ht]
\centering
\includegraphics[width=13cm,trim={3cm 3cm 2cm 3cm},clip]{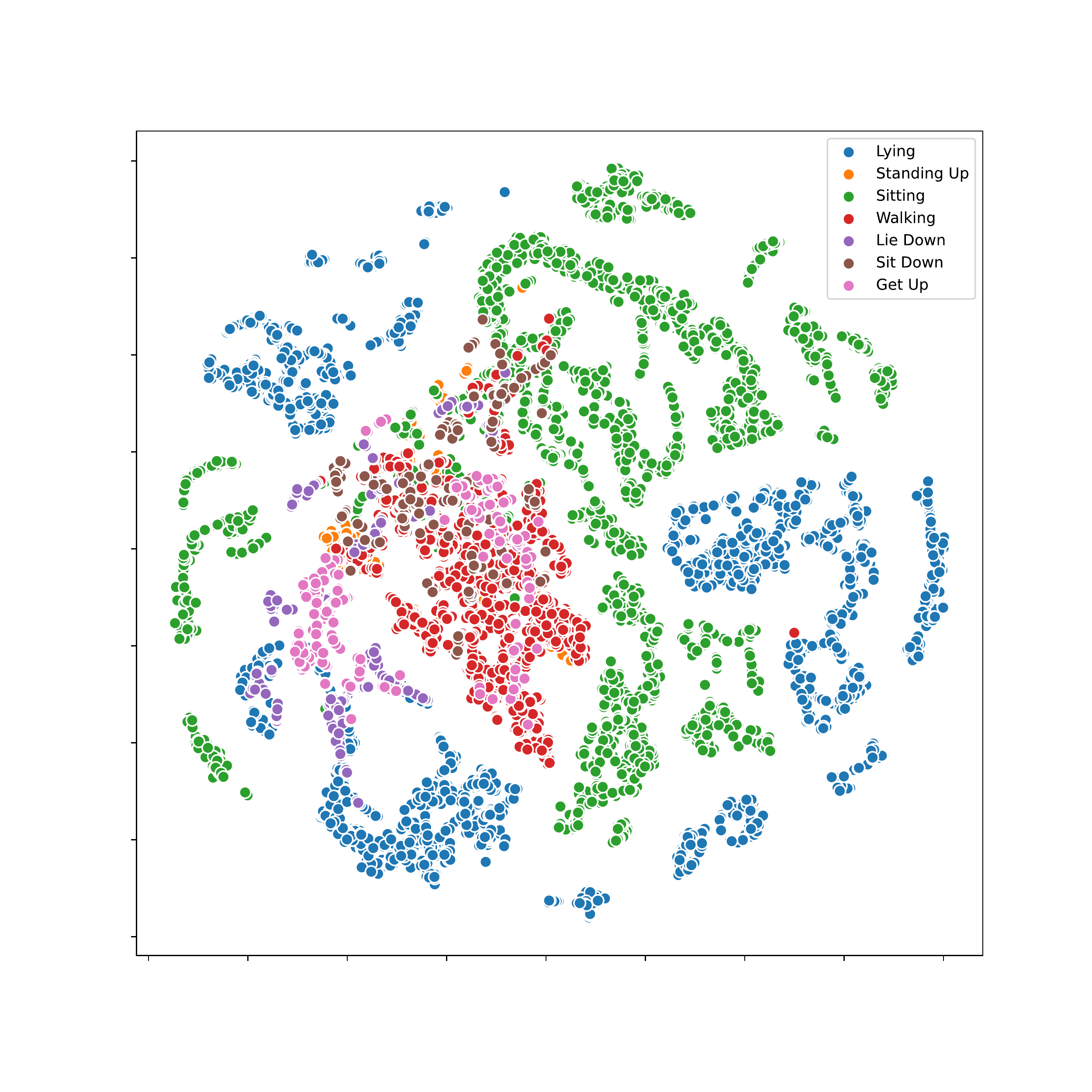}
\caption{\textit{2D visualization of the second last layer output in the test set. (Best viewed in color)}}
\label{fig:t-sne}
\end{figure}

\subsubsection{Visualizing the Feature representations}

In GNN, it is more appropriate for our model to learn vectors for nodes on Euclidean space that are highly comparable to the node's structural representation (representation in graph network). However, in this situation, we are concerned with graph classification problems. To do this, it is essential that all graphs belonging to the same class are well clustered and do not overlap with other class graph representations. To demonstrate our method's performance, we show Figure~\ref{fig:t-sne}. These representations are derived from the network's second last layer. Two neurons are utilized to visualize the sample in the Euclidean space. Notably, it can be seen that actions with similar patterns overlap a tough challenge, providing an opportunity for additional study.

\begin{figure}[ht]
\centering
\includegraphics[width=13cm,trim={0 15cm 0 15cm},clip]{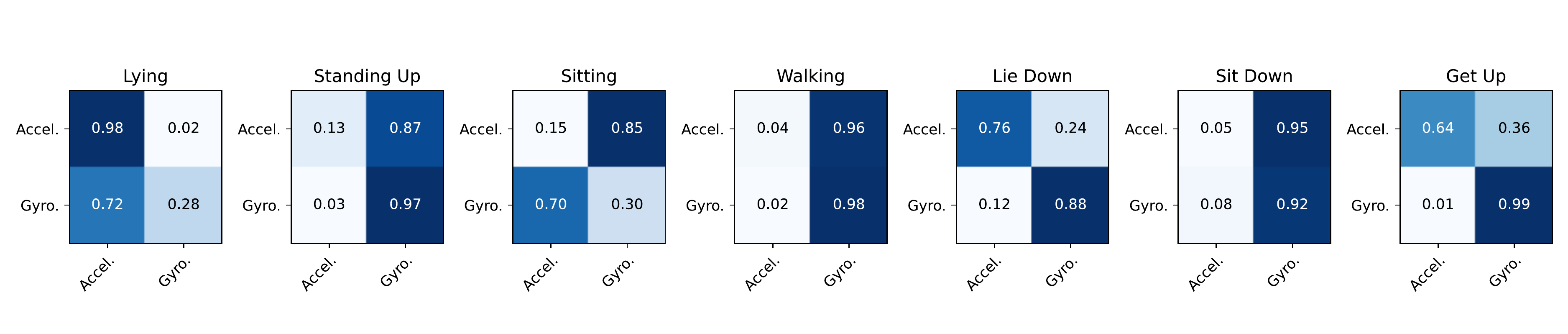}
\caption{\textit{Visualization of attention distribution of different activities in the test set.}}
\label{fig:attention}
\end{figure}

\subsubsection{Inter-Sensor Self-Attention Distribution}

As seen in Table~\ref{tab:baseline}, learning interaction and prioritizing sensors significantly increased our framework's performance. To illustrate, we visualize the learned self-attention feature map for each activity generated by our self-attention encoder (Figure~\ref{fig:attention}). Indeed, sensors do contribute differently to various activities. For example, sensor 1 is more important to the prediction for "lying," both sensors chose to preserve only the representation of sensor 1. For "lie down", both sensors themselves already have enough information, so they keep their representations.

\section{Conclusion}
To enhance the performance of the benchmark HAR methods, we proposed GCN with self-attention in this work. Furthermore, we contributed by developing a time-series-graph module that converts raw HAR data to graphs. We achieved significant improvements on the intricate and challenging dataset (hospital old age patient activities data). The results demonstrate that our combination of a graph structural deep learning framework and a self-attention model overcomes the limitations of conventional deep learning approaches and proves their efficacy by outperforming them. Nonetheless, GNN-based HAR opens up an unlimited number of possibilities, including the exploration of edge features and graph distances. These will be our future development directions, and we believe they will enable us to establish new benchmarks for HAR domain performance. We expect that our study will pave the way for further research into GNN for the HAR domain.

\bibliographystyle{plain} 
\bibliography{main} 

\end{document}